\RequirePackage{snapshot}
\documentclass[journal]{IEEEtran}
\usepackage[T1]{fontenc}

\usepackage{graphicx,overpic}
\usepackage{amsmath}
\usepackage{amssymb}
\usepackage[unicode=true,bookmarks=true] {hyperref}
\hypersetup{pdftitle={Food recognition and recipe analysis: integrating visual content, context and external knowledge}, pdfauthor={Luis Herranz, Weiqing Min, Shuqiang Jiang}}

\usepackage{float,lipsum}
\restylefloat{table}
\usepackage{mdframed}

\usepackage{subfigure}
\usepackage{multirow}
\usepackage{multicol}
\usepackage{afterpage}

\usepackage{xcolor}
\usepackage{soul}

\begin{document}
\title{Food recognition and recipe analysis: integrating visual content, context and external knowledge}
\author{Luis Herranz,~Weiqing Min~and~Shuqiang Jiang
	\thanks{L. Herranz is with the Computer Vision Center. Universitat Autonoma de Barcelona, 08193 Barcelona, Spain, e-mail: lherranz@cvc.uab.es.\protect \\W. Min and S.Jiang are with Key Lab of Intelligent Information Processing of Chinese Academy of Sciences (CAS), Institute of Computing Technology, CAS, Beijing, 100190, China, e-mail: \{minweiqing@ict.ac.cn, shuqiang.jiang\}@vipl.ict.ac.cn.
	}
}
        
\maketitle
\begin{abstract}
The central role of food in our individual and social life, combined with recent technological advances, has motivated a growing interest in applications that help to better monitor dietary habits as well as the exploration and retrieval of food-related information. We review how visual content, context and external knowledge can be integrated effectively into food-oriented applications, with special focus on recipe analysis and retrieval, food recommendation and restaurant context as emerging directions.
\end{abstract}

Food is an essential component of our individual and social life.
Eating habits have direct impact on our health and well-being, while ingredients, flavors and cooking recipes shape specific cuisines that are part of our personal and collective cultural identities.

Recent technological advances such as smartphones equipped with cameras and other rich sensors, pervasive networks and artificial intelligence have powered new uses of technology related with food. For instance, conventional food logging for diet monitoring requires expertise and effort from the user, and is prone to inaccuracies and forgetting. In contrast, an automatic food annotation system could perform automatic analysis, annotation and logging with minimum human intervention. For instance, photos from smartphones are convenient yet powerful entry points to many applications involving recognition, retrieval or recommendation. In this direction, food-oriented social networks and restaurant review services have bloomed, where food enthusiasts (e.g., foodies, gourmets, cooks) connect and share information (e.g., recipes, photos, comments about restaurants). The analysis of this user-contributed data also provides interesting insight to understand eating habits, cuisines and cultures\cite{min2017b}. This collective knowledge can also, in turn, be leveraged by recognition models to improve their accuracy \cite{Xu2015,Chen2016,min2017, salvador2017learning}.

Thus, reliable food analysis from images is essential for these applications. Despite remarkable advances in computer vision, food recognition in the wild still remains a very challenging problem even for humans. We largely rely on contextual and prior information. Similarly, context and prior knowledge can be integrated in automatic food analysis systems. We review some recent works in this emerging direction.

\begin{figure}[]
	\begin{centering}		\subfigure[]{\textsf{\includegraphics[width=0.98\columnwidth]{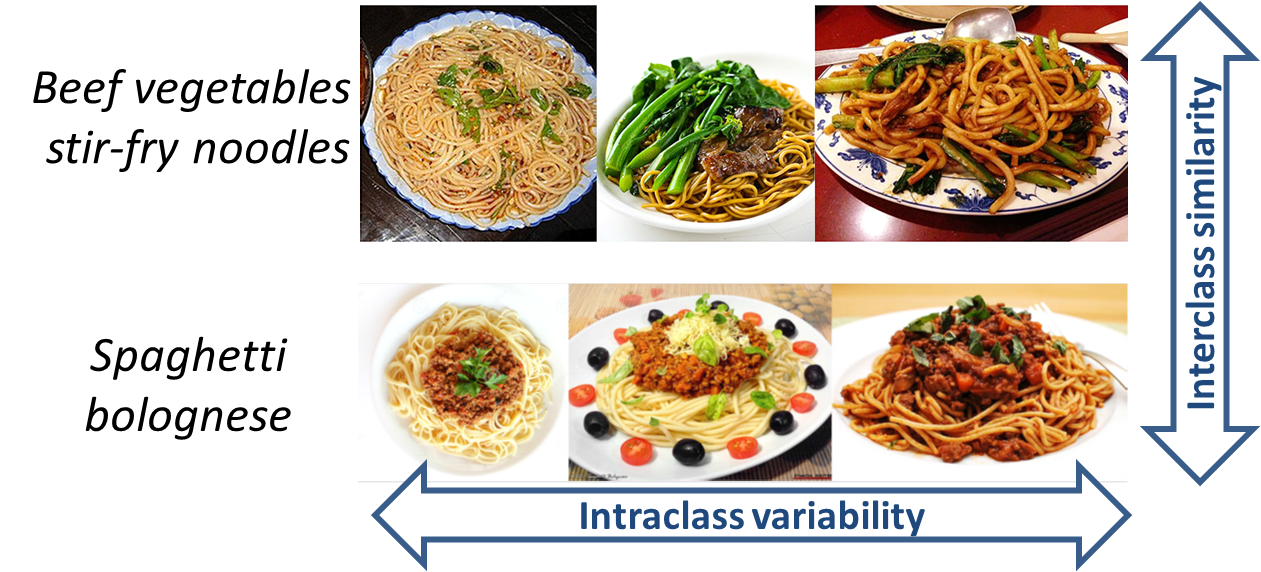}}}
		\par\end{centering}
	\begin{centering}
		\subfigure[]{\textsf{\includegraphics[width=0.98\columnwidth]{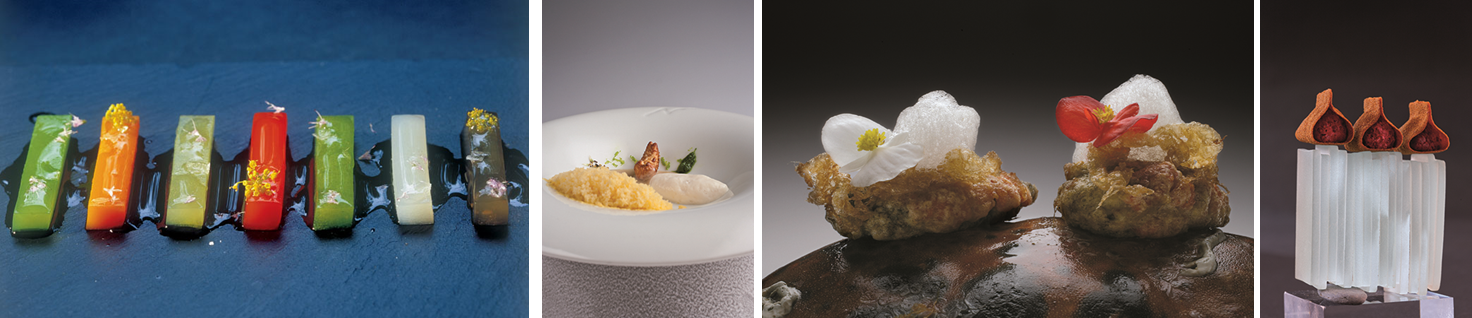}}}
		\par\end{centering}
	\caption{\label{fig:challengingproblem}Examples of dishes (food categories)
		found in restaurants: (a) intraclass variability and interclass similarity,
		and (b) four dishes extracted from elBulli's menu, which includes
		more than one thousand dishes (from left to right): \textit{griddled
			vegetables with charcoal oil}, \textit{curry-flavoured fried brown
			crab mock anemone,} \textit{cashew apple polenta with fermented yoghurt},
		and \textit{azelnut, raspberry and spice bread encerradito}.}
\end{figure}

\section{Challenges in automatic food recognition}

Food recognition can be seen as a particular case of fine-grained visual recognition, where photos within the same category may have significant variability (high intra-class similarity), while are often visually similar to photos from other category (high inter-class similarity, see Figure~\ref{fig:challengingproblem}a). Effective classification requires identifying subtle details and fine-grained analysis.

In the case of restaurants or recipes, the number of categories can explode since the name of dishes in a menu or in a recipe database can be very large. This increases the variability significantly, since the same dish can have very different appearances (due to the particular cooking style, presentation, restaurant, etc). The names also become more elaborated and specific (see Figure~\ref{fig:challengingproblem}b), and many of them are signature dishes that can be found only in one place (e.g., \textit{cashew apple polenta with fermented yoghurt}\footnote{\url{http://www.elbulli.com/catalogo/catalogo/index.ph	p?lang=en}}), for which only one or even no images are available. In general, this constitutes a long tail of rare dishes with very limited training data, which makes the recognition problem from purely visual appearance very difficult.

\begin{figure*}[t]
	\begin{centering}
		\textsf{\includegraphics[width=\textwidth]{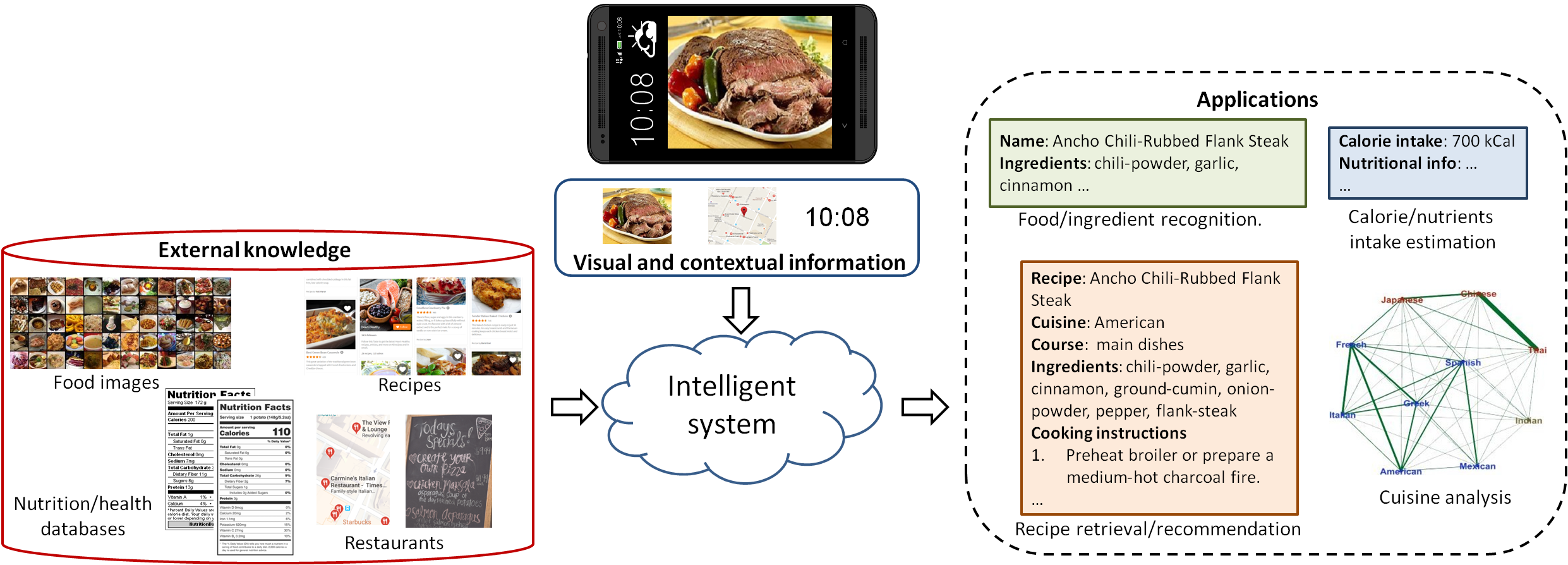}}
		\par\end{centering}
	\caption{\label{fig:overview} Intelligent food analysis system and applications.}
\end{figure*}

\section{Modeling knowledge}
\label{sec:knowledge}

In order to solve complex recognition and understanding problems, humans also analyze the context and rely on prior knowledge to simplify and address the problem effectively.
In general, users will use smartphones that can leverage diverse contextual information and access external knowledge sources. In such scenario, context becomes very important, often more than the content itself.
Figure~\ref{fig:overview} shows a context-aware framework for food applications, where the knowledge about dish names, restaurants, ingredients, nutritional facts, etc. can be exploited by the system together with the particular context of the user.

\subsection{Food visual models}
\label{sec:food_visual_models}
Since images are often the input to our system, a powerful visual model to extract an abstract representation of the image is an important requirement. There is a long history of visual models for general object recognition, from handcrafted descriptors and models to modern data-driven convolutional neural networks (CNNs). 

Similarly, in the first works addressing food recognition, data consisted of a few categories in narrow domains (e.g., fast food). Recognition was address with handcrafted features, pooling and shallow classifiers such as support vector machines (SVMs).

The visual recognition paradigm changed rapidly after the appearance of the ImageNet dataset, with more than one million images, demonstrating the power of data-driven feature learning in the form of deep CNNs. Since then, CNNs have also been the most effective architecture to address visual recognition, and food recognition in particular \cite{kagaya2014food}.

\subsection{Contextual information}
Visual information is just part of all the information available to solve a particular task. In particular, there are many contextual cues humans leverage daily to solve complex tasks\footnote{We use the term \textit{context} for the external context involving the act of picture taking with a smartphone or other device, while for other types of information, intrinsically related with food, we refer to them as \textit{(other) knowledge}. Some works also refer to them as contextual information.}.

Among all the contextual cues that can be considered for food recognition, \textbf{location} is probably the most helpful one. For instance, if the location is Italy, the likelihood of having \textit{spaghetti} is higher than having \textit{noodles}. Location can be considered at multiple scales, from large geographic regions with specific culinary characteristics to smaller scales such as restaurants. Thus, location information can be present in different datasets in heterogeneous formats, from country names to GPS coordinates, restaurant names or street addresses\cite{Xu2015}.

Another very discriminative cue is \textbf{time}. Our daily interaction with food is often based on time-specific meals, e.g., breakfast, lunch, dinner. At a broader scale, the food consumed during working days if often different from weekends or holidays. Seasons also determine which ingredients are available and consequently which dishes are more likely to be observed.

\subsection{Food-specific knowledge}
Any information related with food can also be leveraged for training a better model or for improved inference, while also enabling other applications.

\textbf{Ingredients} are the main components of food, and consequently convey important information for recognition and analysis. Many of them are related with specific flavors and cuisines. Certain ingredients are not directly observable (e.g., salt), but yet useful because of implicit correlations across dishes, flavors and other ingredients\cite{min2017}. They can be seen as attributes and enable extended forms of recognition based on attribute detection and applications based on correlation analysis. Some works also define ingredient-specific attributes, such as cutting and cooking attributes\cite{chen2017cross}, or whether they should be excluded from certain diets (e.g., Vegan, Kosher, Halal)\cite{yang2017yum}. 

\textbf{Recipes} combine ingredients and cooking instructions, often illustrated with images to show the final result and sometimes the cooking process. The quality of the data varies greatly depending on the source. Some recipes are very structured, using closed vocabularies and standard units (e.g., 1 gr. of salt) for ingredients and explicit sequences of instructions \cite{salvador2017learning}, while in other datasets recipes are just textual descriptions described in natural language. A more challenging direction considers cooking videos, adding a temporal dimension to the visual information and with instructions narrated in audio rather than text \cite{malmaud2015s}. Recipes enable popular applications such as recipe retrieval \cite{min2017,salvador2017learning} and recommendation\cite{min2017a}. 

Information about \textbf{restaurants} such as menus, cuisine styles and the location can be very helpful for analysis and context
-based applications. Restaurant information can enable applications such as restaurant recommendation and greatly improve food recognition using the context\cite{Xu2015,herranz2017}.

While not directly helpful for recognition, \textbf{nutritional information} is another type of external information that acts as bridge to many medical and dietary applications such as intake monitoring and diet planning and recommendation. Sometimes dishes in a menu may include explicit calorie or nutritional information. In general, they can be defined at the meal or dish level (e.g., some fast food restaurants include this information) or at the ingredient information \cite{yang2017yum}.

Information about \textbf{users} can be useful in many applications. For instance, profiles, preferences (e.g., preferred food, liked and disliked ingredients, favorite recipes), feedback and comments (usually via sentiment analysis) to improve the performance of recognition and recommender systems \cite{yang2017yum}.

\subsection{Food datasets}

Food datasets with images have proliferated in recent years. Table~\ref{datasets} shows a list of datasets reported in the literature and some of their characteristics. Although some datasets can be used to evaluate multiple tasks, we will roughly distinguish between three groups, according to the main task: general food recognition, recipe analysis/retrieval, and restaurant-based recognition. Large datasets are mostly collected from data in the web, while smaller ones are often captured by the authors.

General \textbf{food recognition} datasets typically consist of images and the corresponding (food) class labels, and they are mainly used to train food classifiers. These datasets have evolved to include progressively more food classes, from early datasets with a few number of cuisine-specific images to larger datasets that include a much larger number of images per class, and cover wider ranges of foods and cuisines. These larger datasets are very suitable for fine tuning deep CNNs leading to state-of-the-art food recognition.

Datasets for \textbf{recipe analysis and retrieval} incorporate ingredients and possibly other cooking information. The number of ingredients can vary from a few tens \cite{Zhou2016} to several thousands \cite{min2017,min2017a,min2017b,Chen2017,salvador2017learning,chen2017cross}. Other interesting attributes that some datasets include are course and cuisine types \cite{min2017,min2017b}, structured cooking instructions \cite{salvador2017learning}, cooking and cutting attributes \cite{chen2017cross} and flavor attributes \cite{min2017a}. They are typically used for cross-modal recipe retrieval, ingredient recognition and cuisine analysis.

Recently, several \textbf{restaurant-centric} datasets have been proposed to evaluate context-based food recognition and logging \cite{Xu2015,Beijbom2015}. In general they are structured as restaurants, a menu with a set of dishes, and photos for these dishes. Xu \textit{et al.} \cite{Xu2015} also included the specific geolocation of the restaurants.

\begin{table*}[htbp]
	\caption{Recent food and recipe datasets. Annotation: \textbf{M}=manual, \textbf{W}=web. Tasks: \textbf{C}=classification, \textbf{R}=retrieval, \textbf{CA}=cuisine analysis, \textbf{IR}=ingredient recognition.}
	\begin{center}
		\begin{tabular}{lccccccccccc}
			\hline
			\multirow{2}{*}{Dataset} & \multicolumn{2}{c}{Basic details}  & \multicolumn{3}{c}{Recipes} & \multicolumn{3}{c}{Restaurants} & \multirow{2}{*}{Annot.} & \multirow{2}{*}{Tasks} \\
			& \#items & \#class & Cuisine & Ingr. & Other & \# & Menu & Geoloc. &  &  & \\  \hline
			UEC Food256\cite{Kawano2014b} & 25088 & 256 & Multiple & - & - & - & - & - & W & C \\
			Food-975\cite{Zhou2016} & 37785 & 975 & Chinese & 51 & - & 6 & Yes & - & M & IR \\
			Yummly-28K\cite{min2017} & 27638 & - & Multiple & 3000 & - & - & - & - & W & R,IR \\
			Yummly-44K\cite{min2017a} & 44K & - & Multiple & 3000 & Flavor & - & - & - & W & R,CA\\
             Yummly-66K\cite{min2017b} & 66K & - & Multiple & 2416 & Course, cuisine & - & - & - & W & R,CA\\
			Vireo-172\cite{Chen2016} & 61139 & 172 & Chinese & 353 & - & - & - & - & W & R \\
			Go cooking\cite{Chen2017} & 61139 & - & Chinese & 5990 & - & - & - & - & W & R \\
			Recipe1M\cite{salvador2017learning} & 1M & - & Multiple & 3000 & Instructions & - & - & - & W & R \\
			MenuMatch\cite{Beijbom2015} & 646 & 41 & Multiple & - & - & 3 & Yes & - & M & R \\
			Dishes (6 cities)\cite{Xu2015} & 117504 & 3832 & Chinese & - & - & 647 & Yes & Yes & W & C \\
			\hline
		\end{tabular}
	\end{center}
	\label{datasets}
\end{table*}

\section{Recipe analysis and retrieval}

\begin{figure*}[t]
	\centering
	\subfigure[Bipartite graph\cite{Zhou2016}]{\label{fig:bipartite}\includegraphics[width=0.2\textwidth]{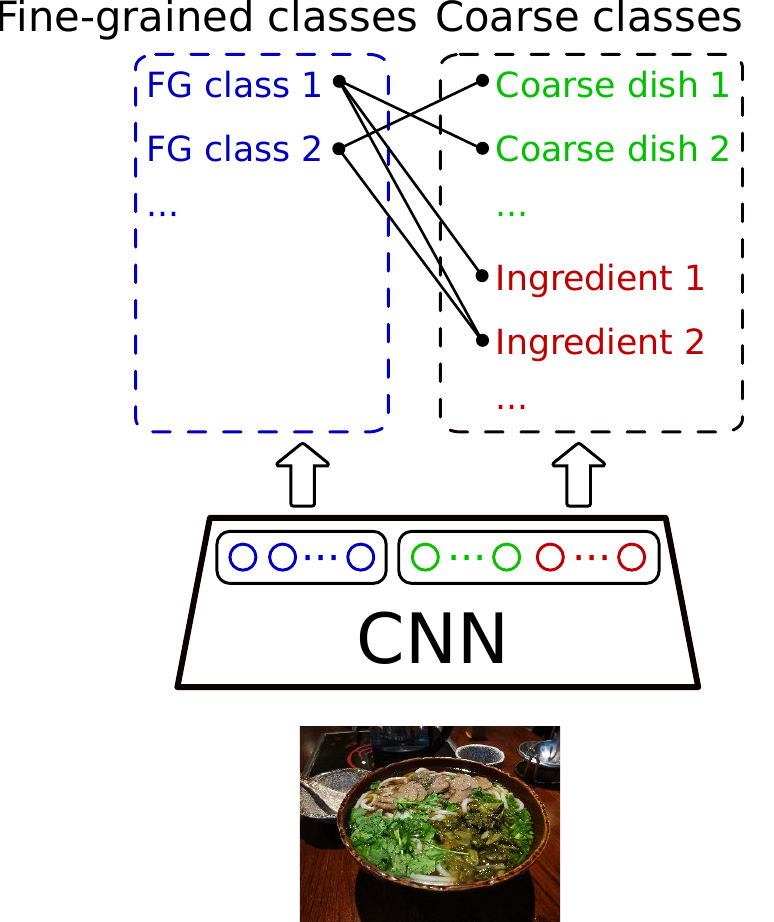}}
    \hspace{0.1\textwidth}
	\subfigure[Deep belief network\cite{min2017}]{\label{fig:m3tdbn}\includegraphics[width=0.2\textwidth]{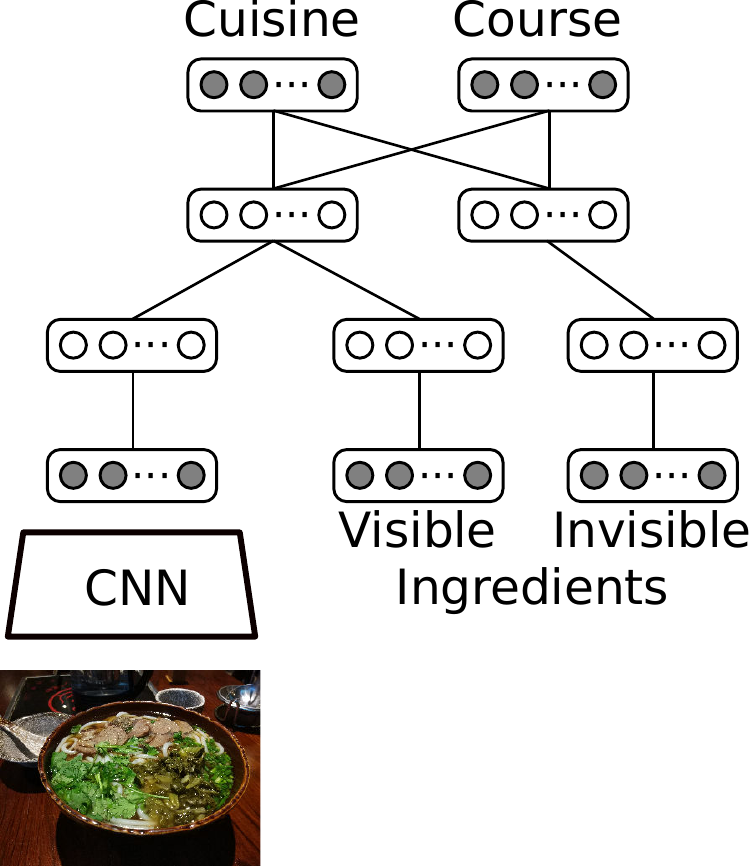}}
    \hspace{0.1\textwidth}
	\subfigure[Encoder-based\cite{salvador2017learning}]{\label{fig:im2recipe}\includegraphics[width=0.2\textwidth]{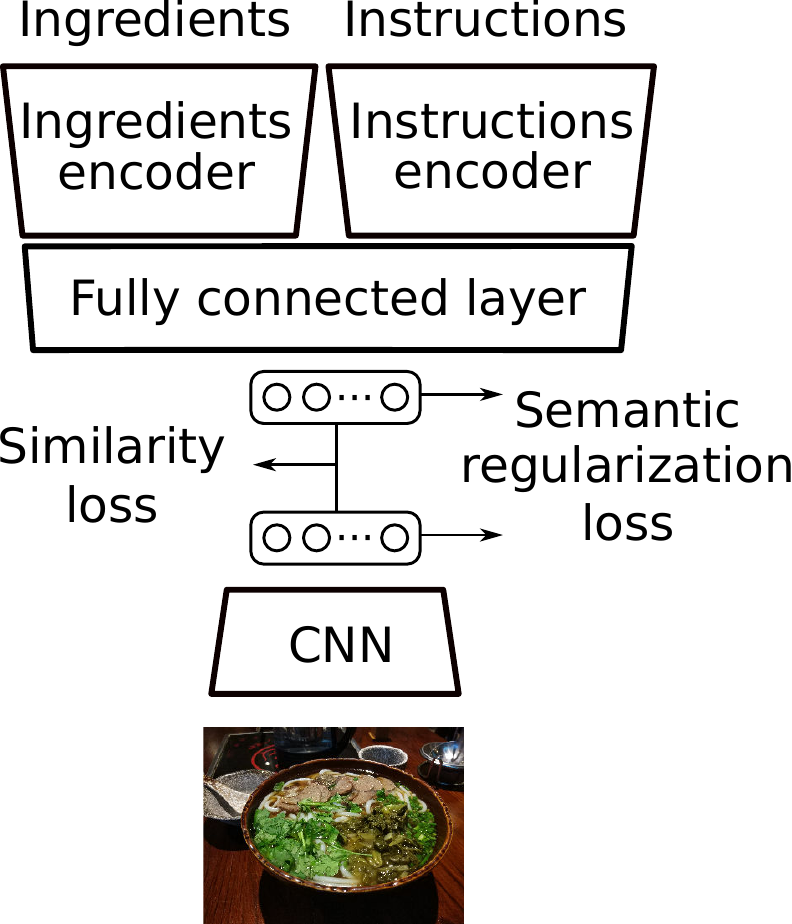}}
	\caption{Multimodal recipe models.}
	\label{fig:recipe_models}
\end{figure*}

Recipes provide valuable information to explore new foods and understand cooking and food consumption habits. In general, ingredient and recipe analysis has focused on textual descriptions, but recent works adopt a multimodal perspective \cite{salvador2017learning,Chen2016,Chen2017,min2017b}, where richer information, such ingredients, cooking instructions, food images and other attribute information are leveraged to better understand recipes and enable novel applications. We review here some of these trends.

\subsection{Cross-modal recipe modeling and retrieval}

Modeling the cross-modal correlation between recipes and images has multiple applications in recognition and retrieval. In most approaches there are two main components: a visual model based on CNNs to encode images, and a recipe encoder. Then a classification or similarity loss is utilized to learn joint representations.

Since food categories and ingredients are closely related, learning to recognize them simultaneously from shared representations can be beneficial for both tasks. Thus, many works use multi-task learning \cite{Zhou2016,Chen2016,min2017}.

As a simple form of recipe, ingredients can be modeled as attributes. Zhou and Lin \cite{Zhou2016} formulate the problem as fine-grained recognition where fine-grained labels are complemented by coarse labels (ingredients and coarser dish classes, respectively). The binary relations between fine-grained classes and coarse classes form a bipartite-graph (see Fig.~\ref{fig:bipartite}). The information of the bipartite graph can be integrated and exploited while training a CNNs.

Chen \textit{et al.} \cite{Chen2016} showed how multi-task learning also enables zero-shot learning and retrieval, that is, new categories where no image has been shown can be recognized by leveraging just a description in terms of ingredients. The authors also showed that additional ingredient-specific information, such as cutting and cooking attributes, can further improve the retrieval and ingredient recognition performance \cite{chen2017cross}.

Min \textit{et al.} \cite{min2017} use deep belief networks (DBNs) to model the correlation between different attributes (e.g., ingredients, cuisine, course) and visual features. Their model explicitly factorizes visible and non-visible ingredients, where the former can exploit more explicit correlations with visual features (see Fig.~\ref{fig:m3tdbn}). Once the model is learned, it can be used to perform inference over the missing variables leading to different applications such as multimodal cuisine classification, cross-modal recipe retrieval and ingredient and attribute inference.

Salvador \textit{et al.} \cite{salvador2017learning} focus on cross-modal recipe retrieval, modeling more complex recipes using recurrent neural networks. Their recipes are structured in two components: a set of ingredients and a list of (textual) instructions (see Fig.~\ref{fig:im2recipe}). Their model encodes recipes as a combination of the representations of two encoders. The first one combines a word embedding followed by an long short-term memory (LSTM) recurrent network to encode ingredients. The second one uses a text encoder to represent each cooking instruction. The resulting representation is fed to another LSTM (note that the cooking instructions are also a sequence) that combines them into a instructions representation. Both representations are combined using a fully connected layer. The recipe and visual representations are aligned during training using a cosine similarity and a semantic regularization loss. Finally, this aligned representation can be used for cross-modal retrieval.

\subsection{Cooking video understanding}
Videos can illustrate recipes and their cooking process better than images. However, the recipe data becomes further complex (temporal dimension) and multimodal (adding audio). This case requires further understanding of actions and speech, and their correct alignment with the corresponding textual instructions.

Malmaud \textit{et al.} \cite{malmaud2015s} investigate these problems combining speech and visual recognition in a hidden Markov model (HMM) framework, and using heterogeneous knowledge collected from different web sources. Results show the effectiveness of their method for recipe text-video alignment, and the application in automatic illustration of recipes and search for events within videos.  

Kojima \textit{et al.} \cite{kojima2015audio} designed a system to addresses multimodal scene understanding for a cooking support robot. Combining CNNs and hierarchical HMMs, the robot is able to recognize cooking events, relate them with the recipe and indicate the future cooking steps to the user.

\subsection{Multi-modal cuisine analysis}
Analyzing the data in recipes shared in the web can provide deep understanding of cultures, regions and individuals, and their relations. Visual features extracted from food images can also be valuable signals in food analysis. For instance, \textit{Yang et al.} \cite{yang2017yum} use food images to learn the food preferences of users. 

Recently, Min \textit{et al.} \cite{min2017b} analyze recipes where ingredients are enriched with images and other attributes such as cuisine and course. Topic models are widely used to learn latent factors that can provide deeper insight about the data. The authors propose a Bayesian Cuisine-Course Topic Model (BC$^{2}$TM) to discover cuisine-course specific topics. Using manifold ranking over the learned distributions, they can retrieve relevant food images for topic visualization, since this method is capable of integrating both deep visual features and semantic topic-ingredient features. This model has a number of applications in visualization, including multi-modal cuisine summarization (i.e. providing the most representative ingredients and images of a given cuisine) and cuisine-course topic analysis and cuisine recommendation.

A recent extension also includes flavors, and models the relation between these textual attributes using a multi-attribute topic model \cite{min2017a}. The resulting topic model is then combined with the visual features in a joint embedding, and applied to flavor analysis and comparison across cuisines, extended cuisine summarization and recipe recommendation.

\section{Food recommender systems}

Recipe representations can be applied directly to cuisine\cite{min2017b} and recipe recommendation\cite{min2017a}. However, more elaborated recommender systems require collecting feedback and user preferences, and in particular, taking health and nutritional aspects in the recommendation. Typically, food recommender systems require a suitable representation of the recipe, nutritional information, personal context, annotations, social context, feedback and external knowledge, usually based on ranking methods. Recently, image has been incorporated as a powerful feature in these systems.

Elsweiler \textit{et al.} \cite{elsweiler2017exploiting} study whether users would select or not healthier replacement foods suggested by a recommender system. The authors use a multimodal recipe dataset including recipe names, images, ingredients, nutritional information and recipe popularity, and predict based on simple classifiers.

Collecting user preferences and feedback about food using traditional textual interfaces is difficult, hampering a wider use of these tools. Recently, Yang \textit{et al.} \cite{yang2017yum} proposed a more intuitive interface based on images for preference elicitation. Their method learn a food image embedding based on multitask learning (classification and metric learning), and the dietary profile is used to re-rank and personalize meal recommendations.

\section{Food recognition in restaurants}
\label{sec:restaurant}

We spend quite significant time in restaurants, cafeterias and dining halls. In those cases there is rich contextual information that can be exploited. The most common types are menu information and geolocation (i.e. location of the images and classes in the geographic space), and can be collected manually \cite{Beijbom2015} or by crawling restaurant websites \cite{Xu2015}).

In this scenario, we want to solve the \textit{geolocalized classification} problem in which for a given pair of visual feature and estimated geolocation we want to estimate the dish class. For comparison, the non-geolocalized problem is addressed with a global classifier that estimates the dish (from the aggregated set of dishes in all restaurants, which could be many). Food recognition in restaurants is also a good scenario to study geolocalized and contextual recognition and the associated problems, such as how to model the context, and how to address the shift between the non-geolocalized train and the geolocalized test distributions.

\subsection{Deterministic context}

\begin{figure*}[t]
	\centering
    \subfigure[Deterministic context \cite{Beijbom2015,Xu2015}\label{fig:rule-based}]{\includegraphics[width=0.5\textwidth]{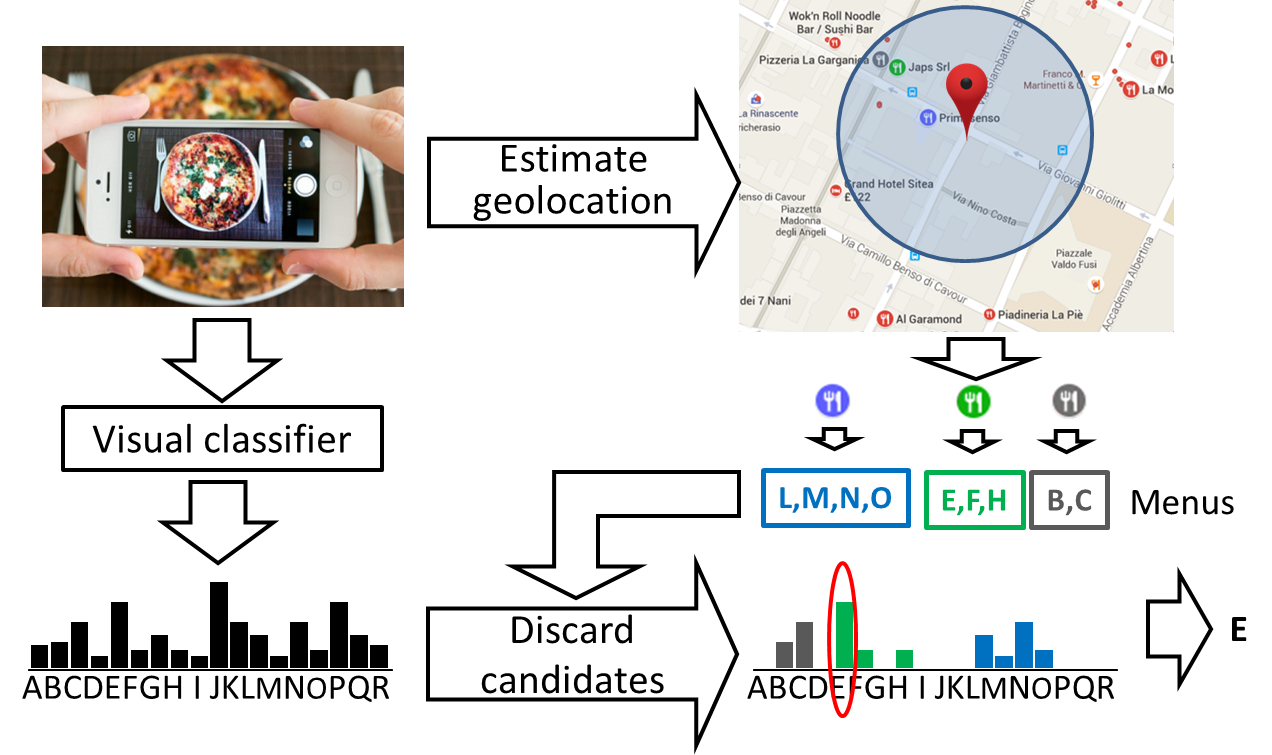}}
    \subfigure[Probabilistic context \cite{herranz2017}\label{fig:probabilistic}]{\includegraphics[width=0.4\textwidth]{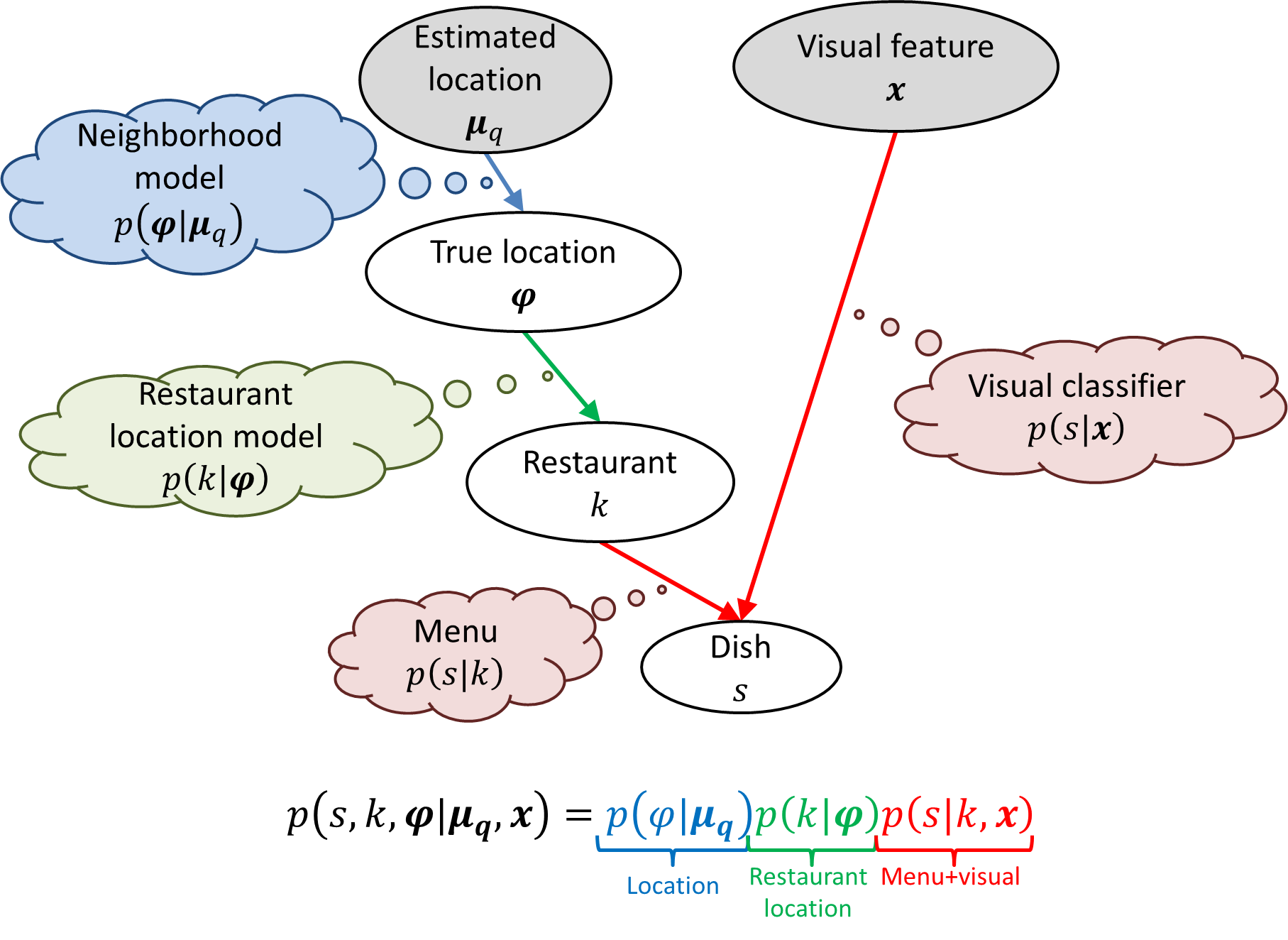}}
	\caption{Food recognition in restaurants.}
\end{figure*}

A common approach consists of using a set of simple deterministic rules to select a set of candidate dishes according to the context (i.e., only those dishes in the restaurant or neighboring restaurants, see Fig.~\ref{fig:rule-based}). In general, either the restaurant is assumed to be known \cite{Beijbom2015} or a few candidate restaurants are selected from those within a fixed distance from the geolocation of the query \cite{Xu2015,herranz2017}. Then the dishes from the candidate restaurants are simply aggregated, and he predictions of the (global) visual for other restaurants are ignored.

Although this simple rule-based approach is simple, it is also very effective and can greatly improve the performance by reducing dramatically the number of relevant classes.

\subsection{Probabilistic context}

A problem with this deterministic context is that they apply a sequence of independent decisions where hard decisions do not take into account uncertainty. In contrast, a probabilistic formulation of the context and the decisions can incorporate more realistic assumptions and take a holistic decision that considers uncertainty.

Herranz \textit{et al.} \cite{herranz2017} reformulate the pipeline in Fig.~\ref{fig:rule-based} as a probabilistic graphical model, where modules are replaced by nodes in the graph, and deterministic rules by probabilistic models (see Fig.~\ref{fig:probabilistic}). The observed variables are the estimated location and the visual feature, both provided by the smartphone. The latent variables are the true location, the restaurant and the dish. The graph introduces explicitly the dependency between the different elements of the problem, which are specified as three submodels: \textit{neighborhood model}, \textit{restaurant location model}  and \textit{(restaurant-conditioned) visual model}. The prediction of the dish is obtained by marginalizing out the other latent variables. 
The rule-based pipeline of Fig.~\ref{fig:rule-based} is a particular case where the restaurant model is a delta (i.e., the restaurant is just a point) and the neighborhood model as just a circular piecewise model (i.e., uniform probability within the a radius and zero outside). However, Gaussian models for location and neighborhood provide better performance. Another advantage of this formulation is that, by performing inference over the other latent variables, we can naturally address related problems such as restaurant recognition and geolocation refinement\cite{herranz2017}.

Probabilistic approaches also allow for more complex models, which could be estimated from data (e.g., data-driven neighborhood models) or from prior information (e.g., restaurant plans, layout data).

\subsection{Geolocalized models}
\begin{figure}
\centering
\includegraphics[width=0.8\columnwidth]{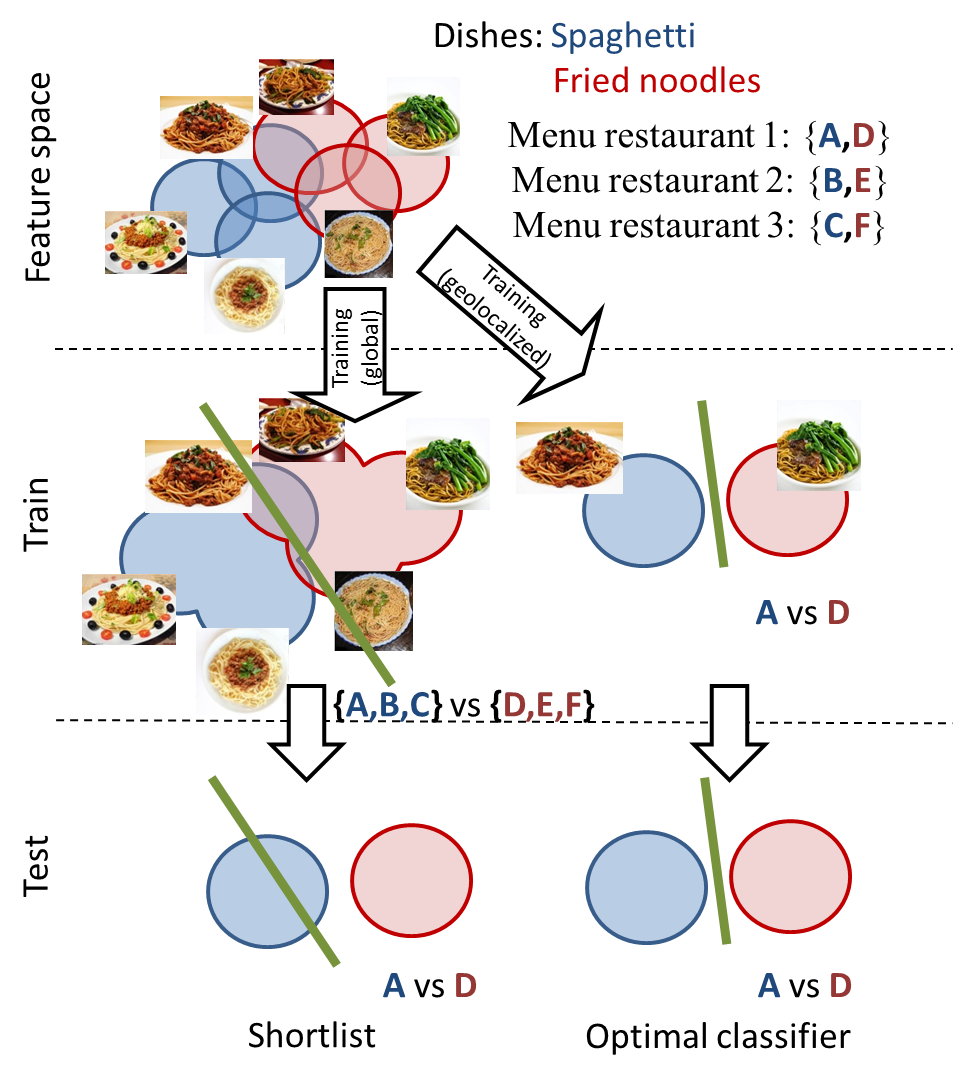}
\caption{\label{fig:geolocalized_models}Global and geolocalized classifiers. After geolocalization only one restaurant is relevant, and thus the problem becomes simpler. However, the global model was trained with the data from all restaurants and the decision boundary is suboptimal.}
\end{figure}

In general, previous systems only apply a posteriori filtering on the results of a global visual classifier\cite{Beijbom2015,herranz2017}. However, the \textit{geolocalized classification} problem becomes simpler during test because only the few categories in the neighboring restaurants are relevant. Since the visual model is global the train and test empirical distributions are different, leading to suboptimal classifiers. The optimal classifier for a particular query would require to train a new classifier with dynamically geolocalized training data, which is not possible in practice (see Fig.~\ref{fig:geolocalized_models}).

Xu \textit{et al.} \cite{Xu2015} proposed exploiting that most images are concentrated in few geographical locations (e.g., restaurants). Using restaurants as anchors, a pool of geolocalized classifiers is trained (with training images from each restaurant and neighboring ones). During test a query-adapted geolocalized classifier is estimated dynamically as a combination of anchor models. The results show that not only the accuracy is higher than using a global visual classifier, but also faster and much more scalable to add new dishes and restaurants.

\vskip 0.2in
Note that both probabilistic and geolocalized models are general approaches not limited to restaurant-based food recognition. They can be applied to similar scenarios, such as shops where many products in a catalog are concentrated in a small geographic area. Furthermore, the same principle could be applied to other factors beyond geolocation (e.g., time).

\subsection{Restaurant-related applications}
In this article we focus on context and knowledge modeling, but there are also interesting applications of automatic food recognition to self-service restaurants and dining halls. For instance, accurate detection and segmentation of the different food items in a food tray can be used for monitoring food intake and nutritional information \cite{Beijbom2015}, and automatic billing to avoid the cashier bottleneck in self-service restaurants \cite{aguilar2017grab}.

\section{Conclusions and future directions}
The integration of multimodal content, context and external knowledge helps human and machines to solve complex problems. In this spirit, we have described a general intelligent framework applied to food analysis, and reviewed some recent advances in several directions, including recipe analyisis, food recommendation, restarant oriented applications and related datasets. A future with even more pervasive intelligent and wearable devices, increasing obesity and cardiovascular diseases and increasing interest in discovering and unverstanding new foods and cuisines suggests research in this area will further develop.

Nevertheless, there are still many open research problems and applications. While visual models have progressed significantly thanks to deep learning, multimodal representations and cross-modal alignments can still be improved. So can be the recognition of categories, ingredient and nutrients, and accurate the estimation of food and nutrients intake. More structured food-related knowledge such as knowledge graphs, together with geolocalized, contextualized and personalized models can also be beneficial in challenging scenarios (e.g., restaurants). We hope this article helps provide an overview of recent research and open problems and outlines some potential research directions.

\bibliographystyle{IEEEtran}
\bibliography{ieeemm_v3}

\end{document}